%% file: main.tex
\documentclass[10pt,twocolumn,letterpaper]{article}

\usepackage{iccv}
\usepackage{times}
\usepackage{epsfig}
\usepackage{graphicx}
\usepackage{amsmath}
\usepackage{amssymb}
\usepackage{booktabs}
\usepackage{multirow}
\usepackage{enumitem}
\usepackage{caption}
\usepackage{indentfirst}
\usepackage{makecell}
\usepackage[accsupp]{axessibility}  

\def\eg{\emph{e.g}.}

\makeatletter

\newcommand{\Rmnum}[1]{\expandafter\@slowromancap\romannumeral #1@}
\newcommand*{\email}[1]{\texttt{#1}}
\makeatother

\usepackage[pagebackref=true,breaklinks=true,letterpaper=true,colorlinks,bookmarks=false]{hyperref}

\iccvfinalcopy

\def\iccvPaperID{6318}

\ificcvfinal\fi
\def\jiang#1{\textcolor{black}{#1}}

\begin{document}
\title{EMR-MSF: Self-Supervised Recurrent Monocular Scene Flow \\ Exploiting Ego-Motion Rigidity}
\author{Zijie Jiang\qquad \qquad Masatoshi Okutomi\\Tokyo Institute of Technology\\
\email{\small zjiang@ok.sc.e.titech.ac.jp, mxo@ctrl.titech.ac.jp}
}
\maketitle
\ificcvfinal\thispagestyle{empty}\fi

\begin{abstract}

    Self-supervised monocular scene flow estimation, aiming to understand both 3D structures and 3D motions from two temporally consecutive monocular images, has received increasing attention for its simple and economical sensor setup. However, the accuracy of current methods suffers from the bottleneck of less-efficient network architecture and lack of motion rigidity for regularization. In this paper, we propose a superior model named \textbf{EMR-MSF} by borrowing the advantages of network architecture design under the scope of supervised learning. We further impose explicit and robust geometric constraints with an elaborately constructed ego-motion aggregation module where a rigidity soft mask is proposed to filter out dynamic regions for stable ego-motion estimation using static regions. Moreover, we propose a motion consistency loss along with a mask regularization loss to fully exploit static regions. Several efficient training strategies are integrated including a gradient detachment technique and an enhanced view synthesis process for better performance. Our proposed method outperforms the previous self-supervised works by a large margin and catches up to the performance of supervised methods. On the KITTI scene flow benchmark, our approach improves the SF-all metric of the state-of-the-art self-supervised monocular method by 44\% and demonstrates superior performance across sub-tasks including depth and visual odometry, amongst other self-supervised single-task or multi-task methods.
\end{abstract}

\input{./Tex/introduction}
\input{./Tex/related_work}
\input{./Tex/method}
\input{./Tex/experiments}
\input{./Tex/conclusion}


{\small
\bibliographystyle{ieee_fullname}
\bibliography{shortstrings,egbib}
}

\end{document}

%% file: Tex/introduction.tex
\section{Introduction}
\label{sec:intro}


{\tabcolsep=1pt
\begin{figure}[t]
    \centering
    \begin{tabular}{cc}
    \includegraphics[width=0.49\linewidth]{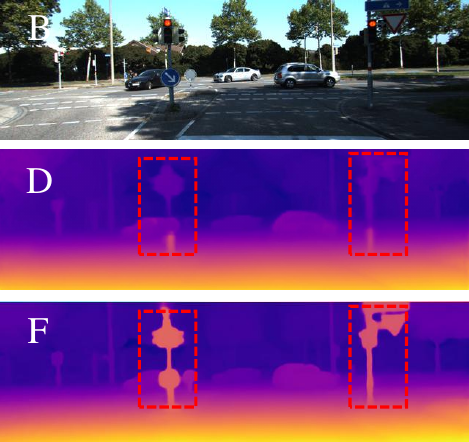} &
    \includegraphics[width=0.49\linewidth]{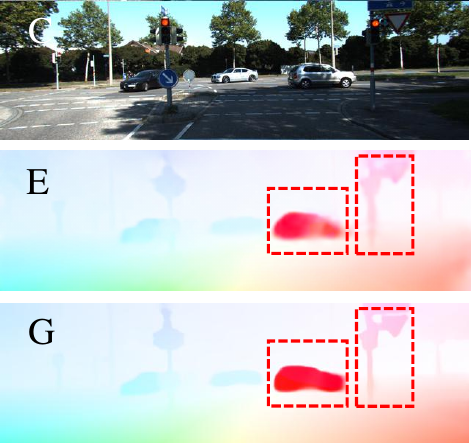} \\
    \end{tabular}
    \caption{{\bf Comparison between our method and \cite{hur2020self}.} (a) input first frame, (b) input second frame, (c) depth of first frame from \cite{hur2020self}, (d) synthesized optical flow  from \cite{hur2020self}, (e) depth of first frame from our method, (f) synthesized optical flow from our method. Our method generates more regularized and detailed predictions as shown in red boxes.}
    \label{fig:compare}
\end{figure}
}

Scene flow estimation, which involves estimating both 3D structure and 3D motion of a dynamic scene from its two consecutive observations, has been receiving increasing attention due to its significance in areas such as robotics \cite{dewan2016rigid}, augmented reality \cite{jaimez2015motion}, and autonomous vehicles \cite{menze2015object}.
Recently, deep learning has demonstrated remarkable progress in the domain of scene flow estimation based on various input modalities, including stereo images \cite{behl2017bounding, jiang2019sense, ma2019deep, schuster2018sceneflowfields, vogel20153d, ren2017cascaded}, RGB-D pairs \cite{lv2018learning, Qiao2018sfnet, teed2021raft, mehl2023m}, or Lidar points \cite{liu2019flownet3d, gu2019hplflownet, wang2020flownet3d++, wu2020pointpwc, puy2020flot, wei2021pv, dong2022exploiting, cheng2022bi, ding2022fh, wang2022matters}.
These methods, however, either require strict sensor calibrations (\eg, stereo-based), or expensive devices (\eg, RGB-D or Lidar-based) for achieving satisfactory performance, which restricts their widespread applications.

On the other hand, monocular scene flow estimation methods \cite{brickwedde2019mono, yang2020upgrading, yin2018geonet, zou2018df, luo2019every, jiao2021effiscene, hur2020self, hur2021self, bayramli2022raft} which only require a monocular camera for obtaining both 3D structure and 3D motion, have been presented as an economical yet effective solution for dynamic 3D perception.
The methods \cite{brickwedde2019mono, yang2020upgrading} combined with supervised learning have yielded promising results,
yet the primary challenge facing them has been the limited availability of ground-truth training data.
To address this limitation, several multi-task methods \cite{yin2018geonet, zou2018df, luo2019every, wang2019unos, jiao2021effiscene} have been proposed to jointly learn the depth, 2D optical flow and camera ego-motion networks from monocular sequences in a self-supervised manner, and the scene flow can be calculated from the outputs.
Recently, \cite{hur2020self, hur2021self, bayramli2022raft} have shown it feasible to train a single network to directly estimate both depth and 3D scene flow from two monocular images and outperform the previous multi-task methods.
These methods typically build upon a standard optical flow pipeline (\eg, PWC-Net \cite{sun2018pwc} or RAFT \cite{teed2020raft}) as basis and adapt it for monocular scene flow.
Despite the notable progress achieved by these methods, their accuracy still lags behind the supervised monocular methods
by a large margin.

In this paper, we propose a novel approach for self-supervised monocular scene flow estimation, which outperforms the previous methods significantly as shown in Fig.\ref{fig:compare}. To introduce explicit 3D geometry-oriented property, we follow the network architecture proposed in the supervised RGB-D method RAFT-3D \cite{teed2021raft} that iteratively refines a dense SE3 motion field for scene flow estimation.
This improvement of architecture compared to previous methods directly improves the performance to a new level, but we argue that it still lacks the usage of \textit{Ego-Motion Rigidity (EMR)}, an important prior that pixels in static regions should have the same SE3 motion as the ego-motion. A novel module named ego-motion aggregation (EMA) is thus proposed to jointly estimate ego-motion as well as a rigidity soft mask from the dense SE3 motion field.
A new motion consistency loss is elaborately designed for constraining motion estimations in static areas represented by the rigidity soft mask.
However, we notice that the network is inclined to select only a small subset of static regions which leads to a rigidity soft mask of low quality. To mitigate this problem, we adopt an efficient mask regularization loss to encourage the network to locate as many static regions as possible. Further performance improvement is attributed to our proposed training strategies including a gradient detachment technique and an improved view synthesis process.

Our main contributions are summarized as follows:
\begin{itemize}
    \item We propose a novel self-supervised monocular scene flow estimation by incorporating 3D geometry-oriented network architecture property and exploiting ego-motion rigidity (EMR-MSF).
    To the best of our knowledge, we are the first method capable of jointly estimating depth, dense SE3 motion field and ego-motion from monocular images, as well as full scene flow derived from them.
    \item We introduce a novel ego-motion aggregation (EMA) module accompanied by a rigidity soft mask to precisely locate static regions for robust and accurate ego-motion estimation.
    \item We propose two new training losses to constrain the motion estimations in static regions, along with two effective training strategies to enhance accuracy as explained in Sec. \ref{sec:self_training}.
    \item We conduct extensive experiments to verify the effectiveness of our proposed method, resulting in a 44\% accuracy boost in the SF-all metric compared to the previous state-of-the-art method on the task of monocular scene flow estimation, as well as superior results in monocular depth and visual odometry.
\end{itemize}

%% file: Tex/related_work.tex
\section{Related Work}

\begin{figure*}[t]
    \centering
    \begin{tabular}{c} 
        \includegraphics[width=0.95\linewidth]{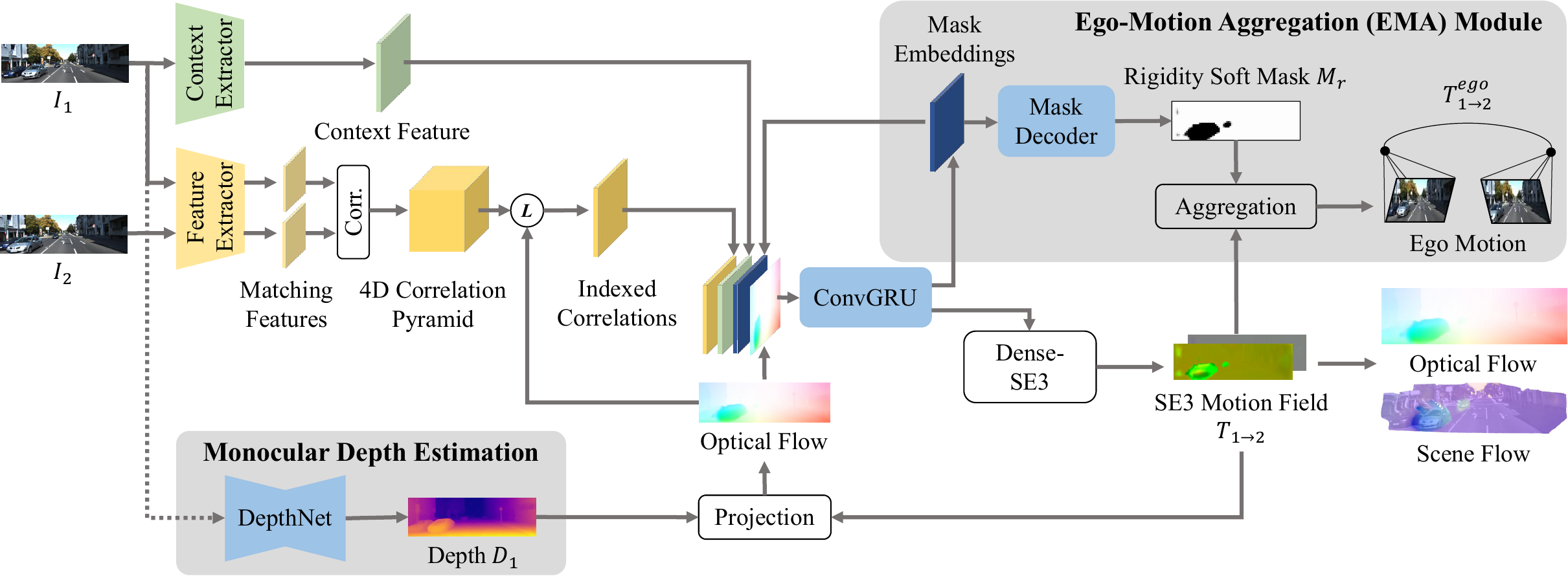}
    \end{tabular}
    \caption{\textbf{Proposed network architecture}.
    We highlight the different parts from RAFT-3D \cite{teed2021raft} with the shaded boxes, including 1) end-to-end trainable monocular depth estimation that substitutes the estimated depths for fixed input depths in the original structure, 
    2) an ego-motion aggregation (EMA) module for inferring ego-motion along with a learnable rigidity soft mask for locating static regions.
    }
    \label{fig:overview}
\end{figure*}

\noindent
\textbf{Scene flow.}
As first introduced in \cite{vedula1999three}, scene flow estimation is defined as the task of jointly estimating 3D structures and 3D motions for each scene point.
The early studies \cite{basha2013multi, huguet2007variational, vogel2013piecewise, vogel2014view, vogel20153d} are based on stereo inputs and approach the scene flow estimation as an energy minimization problem.
Recently, deep learning has demonstrated powerful capabilities in end-to-end learning of scene flow estimation from stereo inputs \cite{jiang2019sense, ma2019deep, schuster2018sceneflowfields}.
Additionally, approaches that leverage pre-existing 3D structure through inputs of RGB-D sequences  \cite{lv2018learning, Qiao2018sfnet, teed2021raft, mehl2023m} or Lidar points \cite{liu2019flownet3d, wu2020pointpwc, puy2020flot, wei2021pv, dong2022exploiting, ding2022fh, wang2022matters} have also been proposed for various scenarios.

\noindent
\textbf{Monocular scene flow.}
The advancement of deep learning techniques has facilitated the acquisition of scene flow solely from monocular images, with early methods relying on supervised learning \cite{brickwedde2019mono, yang2020upgrading}.
To exploit vast amounts of unlabeled data,
a multitude of self-supervised multi-task approaches \cite{yin2018geonet, wang2019unos, zou2018df, luo2019every, jiao2021effiscene, liu2019unsupervised} have been introduced that jointly predict depth, 2D optical flow, and camera motion from monocular sequences.
While the recovery of scene flow is possible using the aforementioned outputs, the accuracy of such estimations is notably inadequate in temporally occluded areas.
Hur et al. \cite{hur2020self} first present a novel self-supervised model capable of inferring depth and 3D motion field from monocular sequences and surpass the performance of previous multi-task methods.
Subsequent studies extend their method into a multi-frame model \cite{hur2021self}, or employ a recurrent network architecture \cite{bayramli2022raft} for better accuracy.

\noindent
\textbf{Rigidity in Scene Flow.}
Scene flow estimation can benefit from prior knowledge about rigidity, which assumes that pixels belonging to the same rigid object should undergo the same rigid transformation.
To leverage the rigidity information in the scene, object detection or segmentation networks are commonly used to identify rigid instances and incorporated in scene flow estimation methods \cite{ma2019deep, cao2019learning, ren2017cascaded, behl2017bounding} for better performance.
Teed et al. \cite{teed2021raft} first propose the rigid-motion embeddings which softly and differentiably group pixels into rigid objects to exploit object-level rigidity.
On the other hand, ego-motion rigidity, where the motion of pixels in static regions is constrained by the camera ego-motion, is widely used in self-supervised multi-task methods \cite{yin2018geonet, wang2019unos, zou2018df, luo2019every, jiao2021effiscene, liu2019unsupervised} but often in a hard and non-differentiable way.
In contrast, our proposed method jointly reasons ego-motion and rigidity soft mask in a fully differentiable manner, providing more robust and accurate scene flow estimation.

%% file: Tex/method.tex
\section{Proposed Method}
Given two temporally consecutive monocular images $\{I_1, I_2\}\in \mathbb{R}^{H\times W\times 3}$, our method aims to recover 1) the corresponding depth maps $D_1, D_2\in \mathbb{R}^{H\times W\times 1}$, 2) the dense SE3 motion field $T_{1\rightarrow 2}\in SE(3)^{H\times W}$ that assigns a rigid transformation to each pixel of $I_1$ to $I_2$, and 3) the ego-motion $T_{1\rightarrow 2}^{ego}\in SE(3)$ from $I_1$ to $I_2$.
The optical flow $F_{1\rightarrow 2}\in \mathbb{R}^{H\times W\times 2}$ and scene flow $S_{1\rightarrow 2}\in \mathbb{R}^{H\times W\times 3}$ from $I_1$ to $I_2$ can be further recovered from the estimated $D_1$ and $T_{1\rightarrow 2}$.
In the following sections, we will begin by providing an overview of the proposed network architecture which incorporates effective designs for 3D estimations from a supervised method (Sec. \ref{sec:overview}).
Afterwards, we provide a detailed description of the proposed ego-motion aggregation (EMA) module that we utilize for estimating ego-motion, as well as a learnable rigidity soft mask for effectively locate static regions (Sec. \ref{sec:ema}).
Finally, we elaborate on our self-supervised training in Sec. \ref{sec:self_training}, which includes novel loss functions designed to fully exploit ego-motion rigidity, as well as improved training strategies.

\subsection{Network Overview}
\label{sec:overview}
Fig. \ref{fig:overview} demonstrates the overview of our network.
We highlight the different parts of our network compared to RAFT-3D \cite{teed2021raft}, which is the basis of our network architecture, inside the shaded boxes.
Our network consists of five stages: 1) monocular depth estimation, 2) feature extraction 3) correlation computing, 4) iterative refinement and 5) ego-motion aggregation.
We first employ a monocular depth network to estimate the depth maps of input images instead of the fixed depths used in the original RAFT-3D structure.
We adopt SDFA-Net \cite{zhou2022self} for depth estimation for its superior performance, which infers disparity from a monocular image under the assumption of a fixed baseline, and further converts the disparity into depth using pre-known focal length and baseline values.
For feature extraction, correlation computing and iterative refinement, we utilize the designs of RAFT-3D, which include the construction of a 4D all-pairs correlation pyramid from extracted features of input images and the use of a ConvGRU unit followed by a Dense-SE3 layer for iterative residual refinement of the SE3 field estimate.
The ego-motion aggregation (EMA) module is employed to further infer ego-motion from the estimated SE3 motion field, which is elaborated on in the next section.
The 3D scene flow and 2D optical flow can be synthesized from estimated depth and SE3 motion field for various applications.

\subsection{Ego-Motion Aggregation}
\label{sec:ema}
As demonstrated in our ablation study \ref{subsec:ablation}, the joint learning of the depth and dense SE3 motion field in the self-supervised scenario can lead to significant ambiguities between the estimations of structure and motion,
\jiang{where the estimated SE3 motions of pixels belonging to the same rigid object, \eg, the static regions, may be inconsistent.}
To mitigate such ambiguities, we incorporate the ego-motion estimation into the joint learning to provide additional constraints \jiang{in static regions}.
We propose to aggregate the ego-motion from the estimated SE3 motion field in contrast to previous multi-task methods \cite{wang2019unos,yin2018geonet,zou2018df}, which utilize a separate network to regress ego-motion from input images.
Furthermore, to handle the dynamic regions which are non-relevant to ego-motion, we introduce a learnable rigidity soft mask to predict per-pixel rigidity, thus locating static regions for stable ego-motion estimation.

Our ego-motion aggregation module proceeds in three steps, as shown in the upper-right corner of Fig. \ref{fig:overview}.
We first incorporate the mask embeddings, a 16-channel feature map initialized to zero values, as new inputs and outputs to the convGRU unit, which is iteratively updated alongside the SE3 motion field.
Next, we decode the mask embeddings using a mask decoder consisting of two convolutional layers and a sigmoid activation layer to obtain the rigidity soft mask \jiang{$M_r$}.
The rigidity soft mask assigns a probability to each pixel, indicating the probability of it belonging to the static region.
In the final step, we derive the ego-motion as an aggregation of estimated SE3 motion field based on the learned rigidity soft mask, which is formulated as:
\begin{equation}
    T_{1\rightarrow 2}^{ego}={\rm Exp}(\frac{\sum M_{r}{\rm Log}(T_{1\rightarrow 2})}{\sum M_{r}}),
\label{eqn_1}
\end{equation}
where ${\rm Log}(\cdot)$ maps SE(3) components to the Lie algebra, and ${\rm Exp}(\cdot)$ performs the inverse operation.

As the ego-motion is differentiably computed from the SE3 motion field, the learning of ego-motion will implicitly impose constraints on the estimation of SE3 motion field.
In the next section, we further combine the self-supervised losses with two new losses utilizing the ego-motion estimation and learned rigidity soft mask to explicitly regularize the motion estimations in static regions.

\subsection{Self-supervised Training}
\label{sec:self_training}

\subsubsection{Self-supervised Loss}
\label{subsec:self_loss}
\indent
To enable self-supervised training, the estimated depth $D_1$ of the first image and the SE3 motion field $T_{1\rightarrow 2}$ are first converted into the scene flow representation $(u, v, \Delta D)$ with known camera intrinsics \cite{mehl2023m}, where $(u, v)$ denotes the standard optical flow $F_{1\rightarrow 2}$, and $\Delta D$ denotes the depth change registered to the first frame $I_1$.
We denote $\overline{D}_1=D_1+\Delta D$, which represents the transformed depth map registered to the first frame.
We obtain the 2D rigid flow $F_{1\rightarrow 2}^{ego}$ in the same manner by replacing $T_{1\rightarrow 2}$ with $T_{1\rightarrow 2}^{ego}$.
The losses for our joint self-supervised learning are introduced as follows:

\noindent
\textbf{Temporal Photometric loss.}
We minimize the photometric differences between the original image and the synthesized images from flow field $F_{1\rightarrow 2}$ and $F_{1\rightarrow 2}^{ego}$, formulated by
\begin{gather}
    L_p=\frac{1}{HW}\sum M_{noc} \odot pe(I_1, w(I_2, F_{1\rightarrow 2})), \\
    L_p^{ego}=\frac{1}{HW}\sum M_{ol} \odot M_{noc} \odot pe(I_1, w(I_2, F_{1\rightarrow 2}^{ego})), \\
    pe(I_a,I_b)=\frac{\alpha}{2}(1-{\rm SSIM}(I_a,I_b))+(1-\alpha)\left| I_a-I_b\right|,
\end{gather}
where $\frac{1}{HW}\sum$ is used for the notation of the mean over all pixels and $\odot$ means element-wise multiplication.
$w(\cdot, \cdot)$ is the view synthesis function with the flow field and $pe(\cdot,\cdot)$ measures the photometric difference between two images.
The occlusion mask $M_{noc}$ is derived from the forward-backward consistency check \cite{meister2018unflow} using $F_{1\rightarrow 2}$ and $F_{2\rightarrow 1}$.
\jiang{We additionally use an outlier mask $M_{ol}$ \cite{jiang2020dipe} for calculating $L_p^{ego}$, which masks out pixels with either large photometric errors mainly resulting from possible occluded or moving regions, or very small photometric errors mainly resulting from textureless regions.}
\jiang{Note that $M_{r}$ is not leveraged here for two reasons: 1) $M_{ol}$ performs more stable than the learned mask $M_r$ in the beginning of training. 2) $M_{ol}$ can better locate pixels which are informative for learning ego-motion estimation.}

\noindent
\textbf{Spatial Photometric Loss.}
To address scale ambiguity in monocular scene flow learning, we utilize stereo samples during training as proposed in previous works \cite{hur2020self, hur2021self, bayramli2022raft}. We use the stereoscopic image synthesis loss utilized in \cite{zhou2022self} to regularize depth estimation on an absolute scale and denote it as $L_d$ in our method.

\noindent
\textbf{Geometric loss.}
To constrain the estimated motion field in 3D space, we exploit the geometric consistency between the transformed depth map $\overline{D}_1$ and estimated $D_2$:
\begin{gather}
    L_g=\frac{1}{HW}\sum M_{noc}\odot ge(\overline{D}_1, w(D_2, F_{1\rightarrow 2})), \label{eqn:5}\\
    ge(D_a, D_b)=\frac{\left| D_a - D_b\right|}{D_a+D_b},
\end{gather}
where $ge(\cdot, \cdot)$ measures the normalized difference \cite{bian2019unsupervised} between two depth maps.

\noindent
\textbf{Smoothness loss.}
The $k$-th order edge-aware smoothness loss function is defined as:
\begin{equation}
    L_s(O)=\frac{1}{HW}\sum \left| \frac{\partial^k O}{\partial x^k} \right| e^{-\beta\left| \frac{\partial I_1}{\partial x} \right|} + \left| \frac{\partial^k O}{\partial y^k} \right| e^{-\beta\left| \frac{\partial I_1}{\partial y} \right|},
\end{equation}
where $O$ is a dense prediction, which can be ${\rm Log}(T_{1\rightarrow 2})$, ${D_1}$ and $F_{1\rightarrow 2}$ in our case. We apply first-order edge-aware smoothness loss to ${\rm Log}(T_{1\rightarrow 2})$ and ${D_1}$, denoted as $L_{s,t}$ and $L_{s,d}$ separately, and apply second-order edge-aware smoothness loss to $F_{1\rightarrow 2}$ as $L_{s,f}$.
The total smoothness loss is calculated as $L_s=\lambda_{st}L_{s,t}+\lambda_{sd}L_{s,d}+\lambda_{sf}L_{s,f}$.

\noindent
\textbf{Motion Consistency Loss.}
To further regularize the SE3 motion field in static regions, we propose to explicitly constrain the motion estimations in these regions to be consistent with the estimated ego-motion, formulated as:
\begin{gather}
    L_c=\frac{1}{HW}\sum M_{r}\odot \lvert {\rm Log}(T_{1\rightarrow 2}) - {\rm Log}(T_{1\rightarrow 2}^{ego})\rvert,
 \end{gather}

{\tabcolsep=1pt
\begin{figure}[t]
    \centering
    \begin{tabular}{ccc}
        {\includegraphics[width=0.33\linewidth]{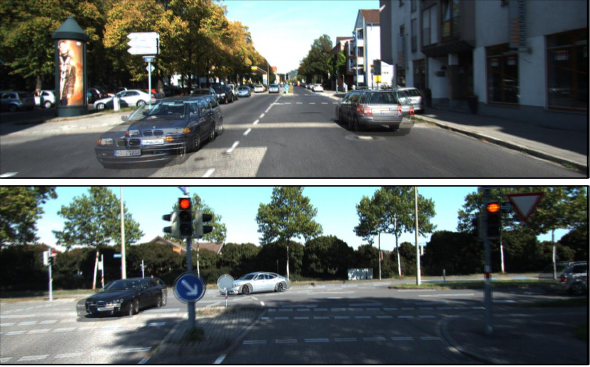}} &
        {\includegraphics[width=0.33\linewidth]{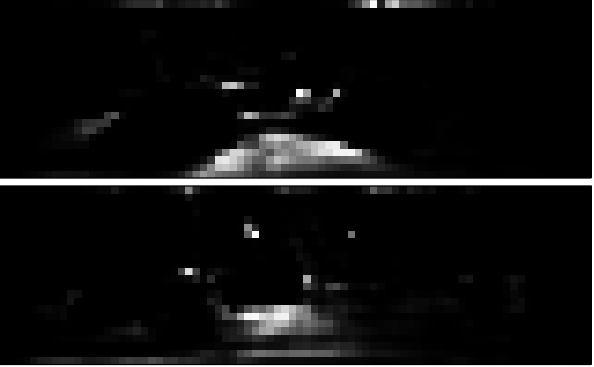}} &
        {\includegraphics[width=0.33\linewidth]{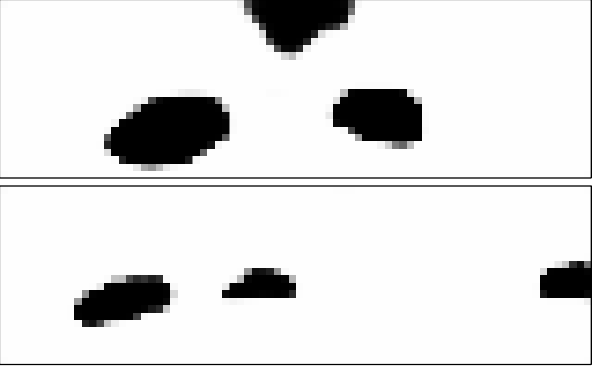}} \\
        Overlayed inputs & W/o $L_{m}$ & W $L_{m}$ \\
    \end{tabular}
    \caption{\textbf{Visualization of estimated rigidity soft masks}. The middle column shows the degeneration cases of estimated rigidity soft mask, which is solved by introducing $L_{m}$ during training.}
    \label{fig_3_mrl}
\end{figure}
}

\noindent
\textbf{Mask Regularization loss.}
We observe that the estimated rigidity soft mask tends to degenerate during training.
This is intuitively reasonable since theoretically the ego-motion can be represented as the SE3 motion of any single pixel in static regions, thus the rigidity soft mask is inclined to select only a small subset of static regions due to $L_c$.
To address this problem, we propose a mask regularization loss to encourage the rigidity soft mask to locate static regions as many as possible for fully exploiting ego-motion rigidity in static regions, which is formulated as:
\begin{equation}
    L_{m}=\frac{1}{HW}\sum\frac{1-M_{r}}{\gamma +M_{r}},
\end{equation}
where $\gamma$ is a hyper-parameter.
We provide a visual comparison of the estimated rigidity soft mask without and with $L_m$ in Fig. \ref{fig_3_mrl}.

\noindent
\textbf{Total Loss.}
We calculate losses for both the final and intermediate estimations from our recurrent structure.
We use an upper-right index $(\cdot)^i$ to denote the losses related to the $i$-th iteration.
The total loss of our method can be summarized as:
\begin{align}
L_{total} &= L_d+\sum_{i=1}^{N}\zeta^{N-i}\left( L_{p}^{i}+ L_{p}^{ego, i}\right. \nonumber\\
& \quad \left. \lambda_g L_{g}^{i}+\lambda_s L_{s}^{i}+\lambda_c L_{c}^{i}+\lambda_m L_{m}^{i} \right),
\end{align}
where $N$ is the iteration number, $\zeta$ is the weight decay factor, and $\lambda=[\lambda_g, \lambda_s, \lambda_c, \lambda_m]$ is the set of hyper-parameters balancing different losses.

\subsubsection{Improved Training Strategies}
\label{subsec:self_strategy}

\noindent
\textbf{Gradient Detachment}.
Our loss functions except $L_d$ are calculated for both the final and intermediate estimations of motion field and ego-motion for preventing divergence of training.
However, joint learning of depth and coarse motion estimations from early iterations can hinder the learning of the depth network.
To address this issue, we propose to detach the gradients of depth estimations when calculating losses using intermediate motion estimations, which ensures that joint learning only occurs when the finest motion estimations are utilized.

\noindent
\textbf{Improved view synthesis process.}
We leverage the full-image warping technique proposed in \cite{stone2021smurf} to provide better supervisory signals at image boundaries during the calculation of photometric loss, which uses cropped images as inputs to the network, but refers to the uncropped images when performing view synthesis.
We further leverage this idea during the calculation of geometric loss in Eqn. \ref{eqn:5}, where we refer to the estimated depths of uncropped images for depth synthesis.

%% file: Tex/experiments.tex
\begin{figure*}[ht!]
    \centering
    \captionsetup{skip=0pt}
    \begin{tabular}{c}
        \includegraphics[width=1\linewidth]{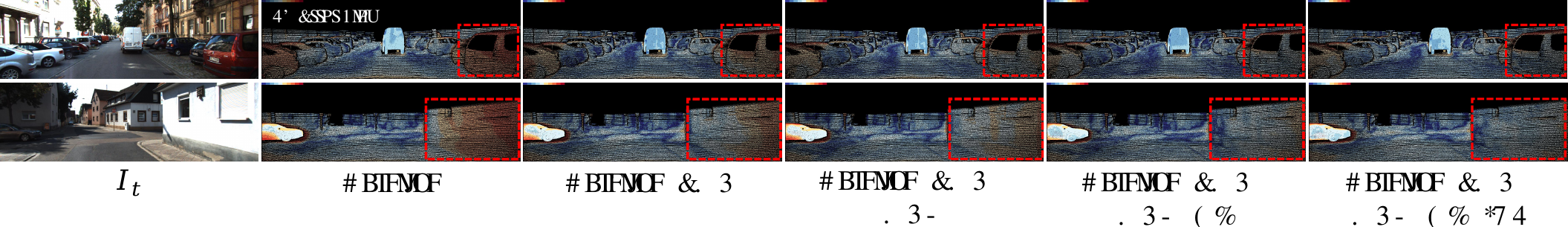}
    \end{tabular}
    \caption{\textbf{Qualitative ablation study of proposed components.} The erroneous predictions are gradually reduced by incorporating proposed components as shown in the red boxes.}
    \label{fig_4_ablation}
\end{figure*}

\begin{table*}[ht!]
\centering
\captionsetup{skip=2pt}
\resizebox{0.75\linewidth}{!}{
\begin{tabular}{cccc|cccc|ccc}
\toprule[2pt]
EMR & MRL & GD & IVS & D1-all $\downarrow$        & D2-all $\downarrow$         & F1-all $\downarrow$         & SF-all $\downarrow$         & EPE-noc $\downarrow$       & EPE-occ $\downarrow$       & EPE-all $\downarrow$       \\ \hline
-   & -   & -  & -  & 13.20         & 21.90          & 14.16          & 28.15          & 2.78          & 12.57         & 4.83          \\
$\surd$   & -   & -  & -  & 9.99          & 18.25          & 13.55          & 24.21          & 2.68          & 10.85         & 4.44          \\
$\surd$   & $\surd$   & -  & -  & 9.83          & 17.25          & 13.65          & 23.07          & 2.69          & 10.17         & 4.23          \\
$\surd$   & $\surd$   & $\surd$  & -  & 9.39          & 16.90          & 13.51          & 22.86          & 2.65          & 10.04         & 4.21          \\
-   & -   & $\surd$  & $\surd$  & 11.73         & 18.76          & 12.54          & 24.80          & 2.74          & 7.63          & 3.81          \\
$\surd$   & $\surd$   & $\surd$  & $\surd$  & \textbf{9.03} & \textbf{15.42} & \textbf{11.93} & \textbf{21.17} & \textbf{2.53} & \textbf{7.07} & \textbf{3.56} \\
\bottomrule[2pt]
\end{tabular}
}
\caption{\textbf{Quantitative ablation study of key components.} EMR: Ego-Motion Rigidity, MRL: Mask Regularization Loss, GD: Gradient Detachment, IVS: Improved View Synthesis. All components effectively improve the performance, especially the EMR component.}
\label{tab_1_ablation}
\end{table*}

\section{Experimental Results}

\begin{table}[ht!]
\centering
\resizebox{\linewidth}{!}{
\begin{tabular}{l|cccc|c}
\toprule[2pt]
Iter. Num. & D1-all $\downarrow$        & D2-all $\downarrow$         & F1-all $\downarrow$         & SF-all $\downarrow$         & Runtime \\ \hline
2          & 9.03          & 15.42          & 11.93          & 21.17          & 127 ms              \\
4          & 8.65          & 13.93          & \textbf{11.36} & 19.05          & 151 ms              \\
8          & 8.38          & 13.14          & 11.76          & 18.31          & 204 ms              \\
12         & \textbf{8.37} & \textbf{12.86} & 11.58          & \textbf{18.11} & 250 ms              \\
\bottomrule[2pt]
\end{tabular}
}
\caption{\textbf{Ablation study of the iteration number.} More iterations give better performance up to about 12, but with a slower speed.}
\label{tab_2_ablation_iter_num}
\end{table}

Our proposed method is evaluated on various tasks including scene flow, monocular depth, and visual odometry.

\subsection{Implementation Details}
\label{subsec:imple_detail}
We implement our network with Pytorch \cite{paszke2019pytorch}.
All components of our network are trained from scratch, except the encoder in the depth network and the context extractor, which use ImageNet \cite{deng2009imagenet} pretrained weights.
We use the Adam optimizer \cite{loshchilov2017decoupled} with $\beta_1=0.5$ and $\beta_2=0.999$ to train our network.
During training, the images are first resized into the resolution of $800\times 240$, and cropped off the top, bottom, left and right 10\% pixels to obtain the input images of $640\times 192$ to leverage the improved view synthesis process.
During test, the images are resized into $640\times 192$ for processing and the results are bilinearly rescaled back to the original size for evaluation.
We use a two-staged training process for better stability of our method.
During the first stage, we separately train the depth network using spatial photometric loss $L_d$ and depth smoothness loss $L_{s,d}$.
Then, we train our full network using the total loss $L_{total}$ for the rest epochs.
The training is carried on for 50 epochs total, 20 epochs for the first stage, and 30 epochs for the second stage.
The initial learning rate is set to 1e-4, and downgraded by half at epoch 20, 25, 30, and 40.
The hyper-parameters of our method are set as:
$[\alpha,\beta,\gamma,\zeta]=[0.15,10,1,0.9], [\lambda_{s,t},\lambda_{s,d},\lambda_{s,f}]=[0.001,1,1], [\lambda_{g},\lambda_{s},\lambda_{c},\lambda_{m}]=[0.1,0.1,0.1,0.1]$, $N=12$.
For data augmentation, we employ random color augmentation, random horizontal flipping and random time order switching.
We use the LieTorch \cite{teed2021tangent} library to perform backpropagation of the SE3 motion field.

\subsection{Datasets and Evaluation Metrics}

\noindent
\textbf{Datasets.}
For the scene flow task, we use the same data setting as previous self-supervised monocular scene flow methods \cite{hur2020self, hur2021self, bayramli2022raft}, which use KITTI Scene Flow Training and Testing as two test sets, and spilt the remaining data into 25801 samples for training and 1684 samples for validation.
For comparison in the task of monocular depth estimation, we follow the data split used in \cite{zhou2022self}, but remove the samples which are the last images of sequences, which gives us 22568 samples for training and 1774 for validation.
The depth evaluation is conducted on the Eigen Test split \cite{eigen2014depth}, which contains 697 images with ground-truth labels.
For the task of visual odometry, we use the official odometry data split, which uses Seq. 00-08 for training and Seq. 09-10 for testing, as done in \cite{wang2019unos, zou2020learning, jiang2022mlfvo}.

\noindent
\textbf{Metrics.}
We follow the evaluation metric of KITTI Scene Flow benchmark \cite{menze2015object} for scene flow estimation, which evaluates the outlier rate of the disparity for the reference frame (D1-all) and for the target image mapped into the reference frame (D2-all), as well as of the optical flow (F1-all).
The outlier rate of the scene flow (SF-all) is obtained by checking if a pixel is an outlier on either of them.
For monocular depth evaluation, we use the publicly used metrics, including: Abs Rel, Sq Rel, RMSE, logRMSE, A1$=\delta\textless 1.25$, A2$=\delta\textless 1.25^2$ and A3$=\delta\textless 1.25^3$.
For visual odometry evaluation, we adopt the KITTI odometry criterion, which reports the average translational error $T_{rel}$ and rotational error $R_{rel}$ of possible sub-sequences of length (100, 200, 800) meters as the main criteria.

\subsection{Ablation Studies}
\label{subsec:ablation}

We first conduct ablation studies to verify the effectiveness of each proposed component of our method on the task of scene flow estimation, including
1) ego-motion rigidity (EMR), which includes the ego-motion aggregation module and losses for $L_{p}^{ego}$ and $L_{c}$,  
2) mask regularization loss $L_r$ (MRL),
3) gradient detachment technique (GD),
and 4) improved view synthesis (IVS).
For efficiency, the ablation studies are conducted using iteration number equal to 2.
We report both the scene flow metrics and end-point-error (EPE) of synthesized optical flow in Tab. \ref{tab_1_ablation}.
Each proposed component proves to be effective in improving the overall scene flow accuracy.
The largest performance gain is obtained by exploiting the ego-motion rigidity, which is in line with our expectation that ego-motion rigidity is an important prior in the task of scene flow estimation.
Fig. \ref{fig_4_ablation} gives a visualization of the achieved error reduction on SF-all error plots from each component.
The erroneous estimations in static regions and image boundaries are largely reduced by incorporating our contributions.
The ablation study on the iteration number is reported in Tab. \ref{tab_2_ablation_iter_num}.
The performance is about to reach convergence when the iteration number is 12.
We also report the runtime for efficiency comparison, which is tested on a single GTX 3090 device for each model.
For the following experiments, we always set the iteration number to 12.

\subsection{Comparison with State of the Art Methods}
\label{subsec:compare}
\noindent


{\tabcolsep=1pt
\begin{figure*}[t]
    \centering
    \captionsetup{skip=-6pt}
    \begin{center}
    \begin{tabular}{cc}
        \rotatebox{90}{\makebox[30pt]{\cite{hur2020self}}} & \includegraphics[width=0.95\linewidth]{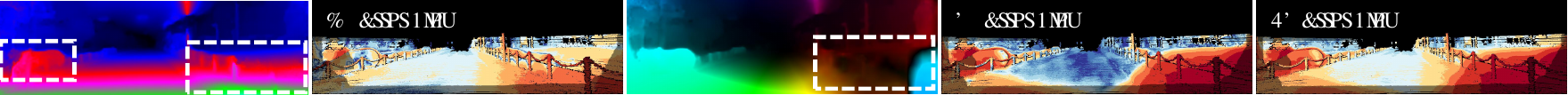} \\ [-3pt]
        \rotatebox{90}{\makebox[30pt]{\cite{hur2021self}}} & \includegraphics[width=0.95\linewidth]{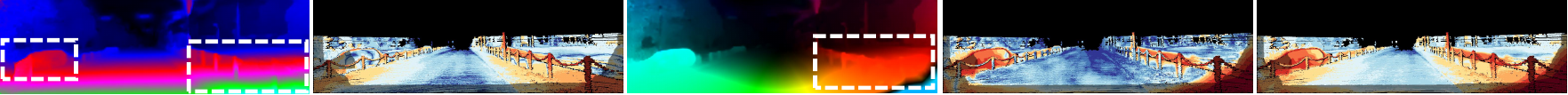} \\ [-3pt]
        \rotatebox{90}{\makebox[30pt]{\cite{bayramli2022raft}}} & \includegraphics[width=0.95\linewidth]{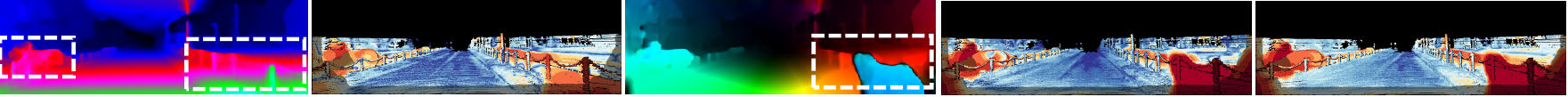} \\ [-3pt]
        \rotatebox{90}{\makebox[30pt]{Ours}} & \includegraphics[width=0.95\linewidth]{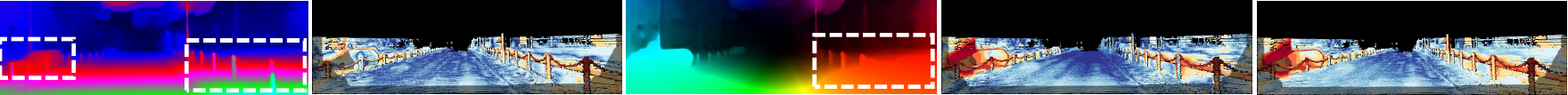} \\ [-0pt]
        \rotatebox{90}{\makebox[30pt]{\cite{hur2020self}}} & \includegraphics[width=0.95\linewidth]{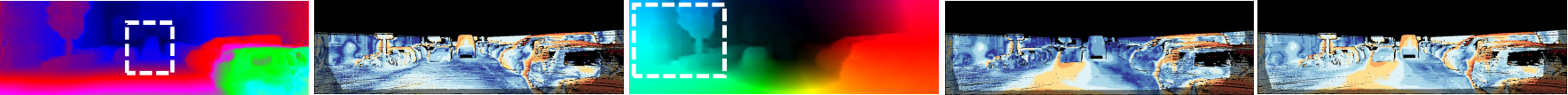} \\ [-3pt]
        \rotatebox{90}{\makebox[30pt]{\cite{hur2021self}}} & \includegraphics[width=0.95\linewidth]{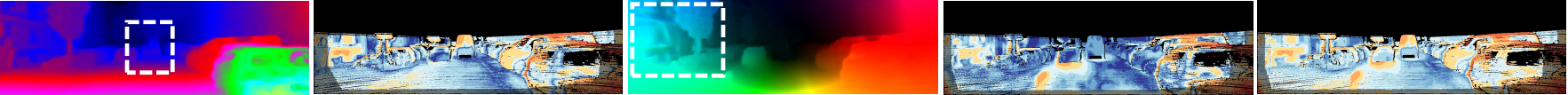} \\ [-3pt]
        \rotatebox{90}{\makebox[30pt]{\cite{bayramli2022raft}}} & \includegraphics[width=0.95\linewidth]{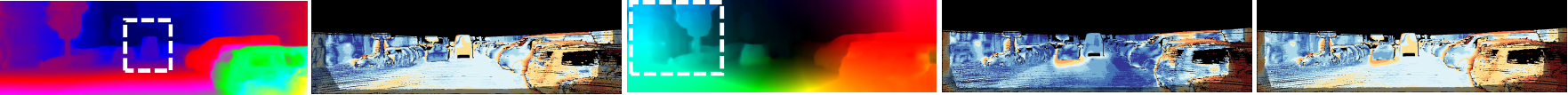} \\ [-3pt]
        \rotatebox{90}{\makebox[30pt]{Ours}} & \includegraphics[width=0.95\linewidth]{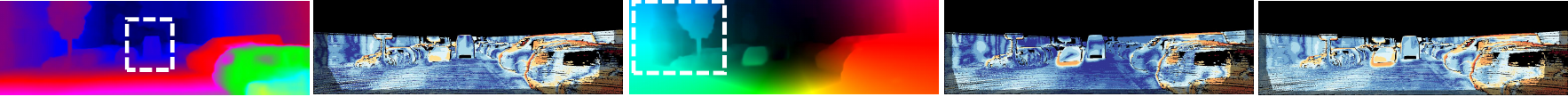} \\
    \end{tabular}
    \end{center}
    \caption{\textbf{Qualitative evaluation on KITTI Scene Flow Testing set.} We compare our method with Self-Mono-SF \cite{hur2020self}, Multi-Mono-SF \cite{hur2021self} and RAFT-MSF \cite{bayramli2022raft} for two scenes using the visualizations provided by the KITTI benchmark \cite{menze2015object}. From left to right: disparity visualization of $I_t$, $D2$ error plot, optical flow visualization, corresponding $F1$ error plot and combined $SF$ error plot.}
    \label{fig_5_comp}
\end{figure*}
}

\begin{table*}[h]
\centering
\renewcommand\arraystretch{1.1}
\resizebox{0.8\linewidth}{!}{
\begin{tabular}{l|cccc|cccc}
\toprule[2pt]
\multirow{2}{*}{Method} & \multicolumn{4}{c|}{KITTI Scene Flow Training Set}               & \multicolumn{4}{c}{KITTI Scene Flow Testing Set}                 \\ \cline{2-9} 
                        & D1-all $\downarrow$        & D2-all $\downarrow$         & F1-all $\downarrow$         & SF-all $\downarrow$         & D1-all $\downarrow$        & D2-all $\downarrow$         & F1-all $\downarrow$         & SF-all $\downarrow$        \\ \hline
Mono-SF \cite{brickwedde2019mono}                 & 16.72         & 18.97          & 11.85          & 21.60          & 16.32         & 19.59          & 12.77          & 23.08          \\ \hline
GeoNet \cite{yin2018geonet}                  & 49.54         & 58.17          & 37.83          & 71.32          & -             & -              & -              & -              \\
DF-Net \cite{zou2018df}                  & 46.50         & 61.54          & 27.47          & 73.30          & -             & -              & -              & -              \\
EPC++ \cite{luo2019every}                   & 23.84         & 60.32          & 19.64          & -              & -             & -              & -              & -              \\
Self-Mono-SF \cite{hur2020self}            & 31.25         & 34.86          & 23.49          & 47.05          & 34.02         & 36.34          & 23.54          & 49.54          \\
Multi-Mono-SF \cite{hur2021self}           & 27.33         & 30.44          & 18.92          & 39.82          & 30.78         & 34.41          & 19.54          & 44.04          \\
RAFT-MSF \cite{bayramli2022raft}                & 18.34         & 23.65          & 17.51          & 30.97          & 21.21         & 27.51          & 18.37          & 34.98          \\
\textbf{EMR-MSF} (Ours)          & \textbf{8.37} & \textbf{12.86} & \textbf{11.58} & \textbf{18.11} & \textbf{9.70} & \textbf{14.51} & \textbf{11.93} & \textbf{19.74} \\
\bottomrule[2pt]
\end{tabular}
}
\caption{\textbf{Quantitative evaluation of the scene flow on the KITTI Scene Flow Training set and Testing set.} The best results are in \textbf{bold}.}
\label{tab:scene_flow}
\end{table*}

\noindent
\textbf{Scene Flow.}
We compare our method with other state-of-the-art monocular scene flow methods on both KTTII Scene Flow Training set and Testing Set as shown in Tab. \ref{tab:scene_flow}.
Our method achieves the best performance among all methods based on self-supervised learning, and even outperforms Mono-SF \cite{brickwedde2019mono}, which is a hybrid method based on the combination of supervised monocular depth estimation and energy minimization.
In Fig. \ref{fig_5_comp}, we visualize the estimations and error maps of our method and other methods on samples from KITTI Scene Flow Testing set.
In the highlighted regions, our method shows better regularized and detailed estimations compare to other methods which give no consideration to exploit ego-motion rigidity.
The error maps of various metrics are provided for better visualization.

\begin{table*}[h]
\centering
\renewcommand\arraystretch{1.1}
\resizebox{0.75\linewidth}{!}{
\begin{tabular}{l|c|cccc|ccc}
\toprule[2pt]
Method       & Sup.                      & Abs Rel $\downarrow$       & Sq Rel $\downarrow$        & RMSE $\downarrow$          & RMSE log $\downarrow$      & A1 $\uparrow$            & A2 $\uparrow$            & A3 $\uparrow$            \\ \hline
Monodepth2   \cite{godard2019digging}       & S             & 0.109          & 0.873          & 4.960          & 0.209          & 0.864          & 0.948          & 0.975          \\
FAL-Net      \cite{gonzalezbello2020forget} & S            & 0.097          & 0.590          & 3.991          & 0.177          & 0.893          & 0.966          & 0.984          \\
PLADE-Net    \cite{gonzalez2021plade}       & S            & 0.092          & 0.626          & 4.046          & 0.175          & 0.896          & 0.965          & 0.984          \\
SDFA-Net     \cite{zhou2022self}            & S            & \textbf{0.090} & \textbf{0.538} & \textbf{3.896} & \textbf{0.169} & \textbf{0.906} & \textbf{0.969} & \textbf{0.985} \\ \hline
EPC++        \cite{luo2019every}            & MS            & 0.127          & 0.936          & 5.008          & 0.209          & 0.841          & 0.946          & 0.979          \\
Self-Mono-SF \cite{hur2020self}             & MS            & 0.125          & 0.978          & 4.877          & 0.208          & 0.851          & 0.950          & 0.978          \\
Monodepth2   \cite{godard2019digging}       & MS            & 0.106          & 0.818          & 4.750          & 0.196          & 0.874          & 0.957          & 0.979          \\
DIFFNet      \cite{zhou_diffnet}            & MS            & 0.101          & 0.749          & 4.445          & 0.179          & 0.898          & 0.965          & 0.983          \\
RAFT-MSF     \cite{bayramli2022raft}        & MS            & 0.093          & 0.781          & 4.321          & 0.186          & 0.901 & 0.960          & 0.981          \\
\textbf{EMR-MSF} (Ours)                              & MS            & \textbf{0.088} & \textbf{0.552} & \textbf{3.946} & \textbf{0.169} & \textbf{0.905}          & \textbf{0.970} & \textbf{0.986} \\
\bottomrule[2pt]
\end{tabular}
}
\caption{\textbf{Quantitative evaluation of the monocular depth on the KITTI Eigen split.} S: trained on stereo pairs. MS: trained on stereo videos. The best results are in \textbf{bold}.}
\label{tab_4_depth_eigen}
\end{table*}

\noindent
\textbf{Monocular Depth.}
We compare our method trained on the KITTI Eigen split with other state-of-the-art monocular depth methods as shown in Tab. \ref{tab_4_depth_eigen}.
Our method achieves the best performance in 4 metrics among all compared methods and second best in the left 3 metrics.
A visual comparison between our results and \cite{zhou2022self} is given in Fig. \ref{fig:depth_comp}.
Our method produces smoother depth estimations than \cite{zhou2022self}, which we attributes to the jointly learning of depth and motion.


{\tabcolsep=0pt
\begin{figure}[t]
    \centering
    \begin{tabular}{ccc}
        {\includegraphics[width=0.33\linewidth]{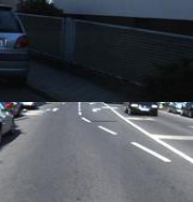}} &
        {\includegraphics[width=0.33\linewidth]{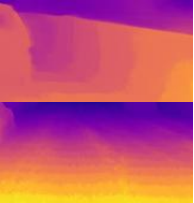}} &
        {\includegraphics[width=0.33\linewidth]{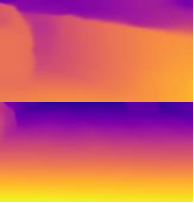}} \\
        RGB & SDFA-Net \cite{zhou2022self} & EMR-MSF (Ours) \\
    \end{tabular}
    \caption{\textbf{Visualization of estimated depth.} We compare our results with SDFA-Net \cite{zhou2022self}.}
    \label{fig:depth_comp}
\end{figure}
}

\begin{table}[h]
\Huge
\centering
\renewcommand\arraystretch{1.1}
\resizebox{\linewidth}{!}{
\begin{tabular}{l|cc|cc}
\toprule[2pt]
\multirow{3}{*}{Method}    & \multicolumn{2}{c|}{Seq.09}   & \multicolumn{2}{c}{Seq.10}    \\
                           & $t_{err}$         & $r_{err}$       & $t_{err}$        & $r_{err}$ \\
                           & (\%) $\downarrow$ & $(^{\circ}/100)$ $\downarrow$ & (\%) $\downarrow$ & $(^{\circ}/100)$ $\downarrow$\\ \hline
ORB-SLAM2 (w/o LC) \cite{mur2017orb}                     & 10.03         & 0.29         & 3.64         & 0.32          \\
ORB-SLAM2 (w LC) \cite{mur2017orb}                     & 3.48         & 0.39         & 3.46         & 0.38          \\ \hline
GeoNet \cite{yin2018geonet}                     & 39.43         & 14.30         & 28.99         & 8.85          \\
Monodepth2 \cite{godard2019digging}             & 17.22         & 3.86          & 11.72         & 5.35          \\
EPC++ \cite{luo2019every}                       & 8.84          & 3.34          & 8.86          & 3.18          \\
LTMVO \cite{zou2020learning}                    & 3.49          & 1.00          & 5.81          & 1.80          \\
MLF-VO \cite{jiang2022mlfvo}                    & 3.90          & 1.41          & 4.88          & 1.38          \\
\textbf{EMR-MSF} (Ours)                                            & 3.49          & 0.78          & 3.11          & 1.04          \\
\textbf{EMR-MSF} (Ours, aligned)                                  & \textbf{3.30} & \textbf{0.78} & \textbf{2.35} & \textbf{1.04} \\
\bottomrule[2pt]
\end{tabular}
}
\caption{\textbf{Quantitative evaluation of the visual odometry.} The best results are highlighted by \textbf{bold} style.}
\label{tab_vo}
\end{table}

\begin{figure}[t]
    \centering
    \begin{tabular}{cc}
        \includegraphics[width=0.45\linewidth]{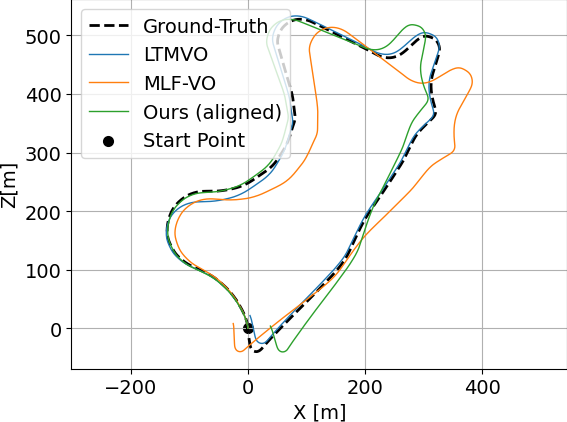} & 
        \includegraphics[width=0.45\linewidth]{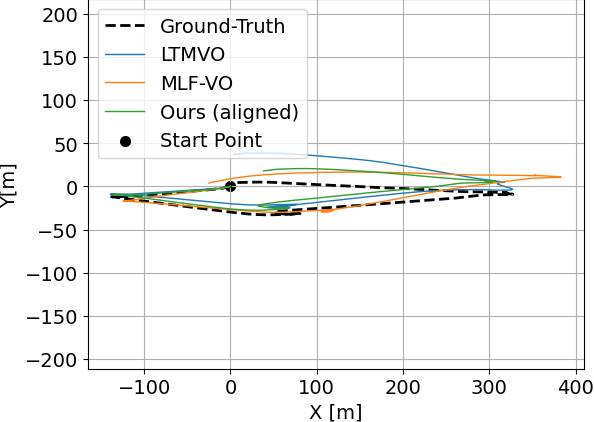} \\
        (a) Top view &
        (b) Front view \\
    \end{tabular}
    \caption{\textbf{Trajectories on Sequence 09 of KITTI Odometry benchmark}. Both the top view and front view are provided for better visualization.}
    \label{fig_traj}
\end{figure}

\noindent
\textbf{Visual Odometry.}
Finally, we compare the performance of our method trained on the KITTI Odometry split with other monocular methods in the task of visual odometry, including ORB-SLAM2 \cite{mur2017orb}, a traditional method, as well as other self-supervised learning-based methods.
We provide both results of ORB-SLAM2 with and without loop closure.
For evaluating monocular methods, we perform the scale alignment to align the predicted up-to-scale trajectories to the ground-truth associated poses using \cite{umeyama1991least}.
Since our method leverages stereo samples during training, it is possible for our method to predict trajectories on a real scale.
For a fair comparison, we provide both aligned and not aligned trajectories of our method in the table.
As shown in Table \ref{tab_vo}, our method outperforms the previous self-supervised learning-based methods in all metrics, and even achieves better accuracy than traditional methods with loop closure in terms of the $t_{err}$ metric.
This demonstrates the effectiveness of our ego-motion aggregation module in improving the accuracy of visual odometry.
We also provide a qualitative comparison of the estimated trajectories from our method, LTMVO \cite{zou2020learning}, and MLF-VO \cite{jiang2022mlfvo} in Fig. \ref{fig_traj}.
Our method yields trajectories with overall smaller drifts than the other methods.

\subsection{Generalization Ability}

\begin{figure}[t]
    \centering
    \begin{tabular}{c}
        \includegraphics[width=0.95\linewidth]{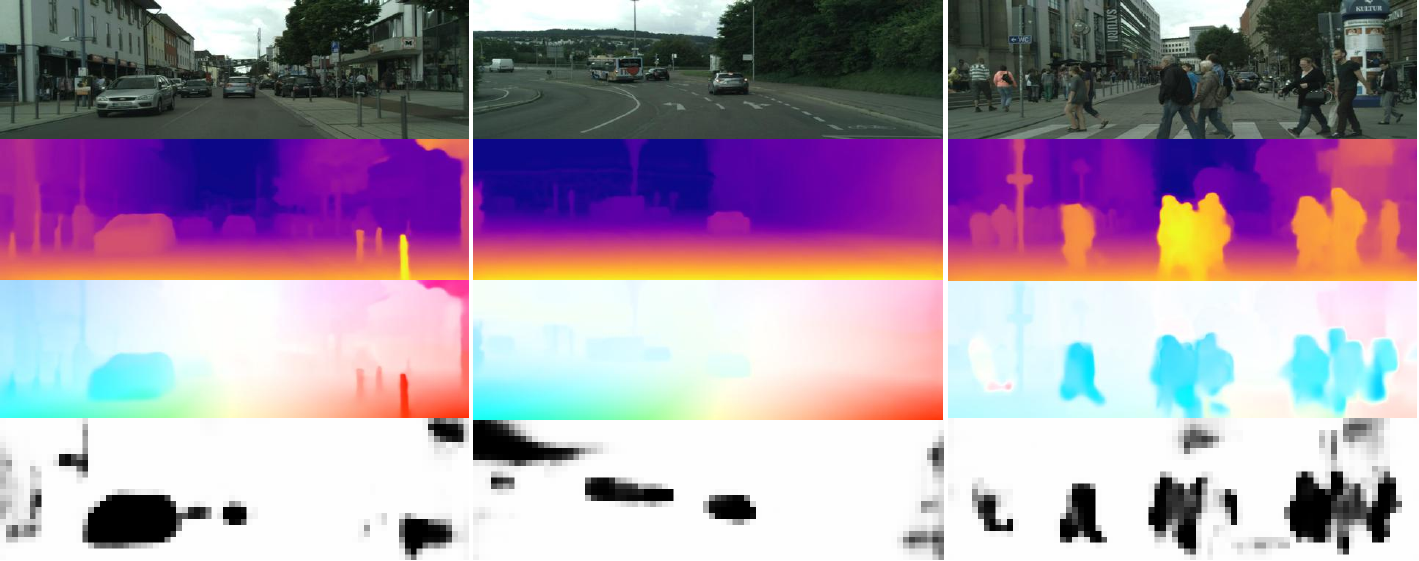}
    \end{tabular}
    \caption{\textbf{Generalization test on Cityscapes \cite{cordts2016cityscapes}.} From top to bottom: input first frame, estimated depth of first frame, synthesized optical flow, estimated rigidity soft mask.}
    \label{fig_gene}
\end{figure}

We use the Cityscapes dataset \cite{cordts2016cityscapes} to test the generalization ability of our model trained on the KITTI dataset \cite{geiger2013vision}.
Several visual samples are provided in Fig. \ref{fig_gene}.
\jiang{Our method remarkably generalizes to unseen data, including some significantly dynamic scenes which are rarely present in the training data, such as the presence of numerous pedestrians crossing before the vehicle.}
More generalization examples can be found in our supplementary material.

%% file: Tex/conclusion.tex
\section{Conclusions}
In this paper, we have proposed a novel self-supervised monocular method named EMR-MSF for scene flow estimation.
Our methods incorporates a 3D geometry-oriented network architecture with novel designs to exploit ego-motion rigidity, which results in well-regularized scene flow estimations from solely monocular images.
Our proposed approach demonstrates promising potential for monocular dynamic 3D perception and is capable of various computer tasks including scene flow, optical flow, depth, and ego-motion estimation.

%% file: main.bbl
\begin{thebibliography}{10}\itemsep=-1pt

\bibitem{basha2013multi}
Tali Basha, Yael Moses, and Nahum Kiryati.
\newblock Multi-view scene flow estimation: A view centered variational
  approach.
\newblock {\em IJCV}, 101:6--21, 2013.

\bibitem{bayramli2022raft}
Bayram Bayramli, Junhwa Hur, and Hongtao Lu.
\newblock Raft-msf: Self-supervised monocular scene flow using recurrent
  optimizer.
\newblock {\em arXiv preprint arXiv:2205.01568}, 2022.

\bibitem{behl2017bounding}
Aseem Behl, Omid Hosseini~Jafari, Siva Karthik~Mustikovela, Hassan Abu~Alhaija,
  Carsten Rother, and Andreas Geiger.
\newblock Bounding boxes, segmentations and object coordinates: How important
  is recognition for 3d scene flow estimation in autonomous driving scenarios?
\newblock In {\em Proc. CVPR}, 2017.

\bibitem{bian2019unsupervised}
Jiawang Bian, Zhichao Li, Naiyan Wang, Huangying Zhan, Chunhua Shen, Ming-Ming
  Cheng, and Ian Reid.
\newblock Unsupervised scale-consistent depth and ego-motion learning from
  monocular video.
\newblock {\em NIPS}, 2019.

\bibitem{brickwedde2019mono}
Fabian Brickwedde, Steffen Abraham, and Rudolf Mester.
\newblock Mono-sf: Multi-view geometry meets single-view depth for monocular
  scene flow estimation of dynamic traffic scenes.
\newblock In {\em Proc. ICCV}, 2019.

\bibitem{cao2019learning}
Zhe Cao, Abhishek Kar, Christian Hane, and Jitendra Malik.
\newblock Learning independent object motion from unlabelled stereoscopic
  videos.
\newblock In {\em Proc. CVPR}, 2019.

\bibitem{cheng2022bi}
Wencan Cheng and Jong~Hwan Ko.
\newblock Bi-pointflownet: Bidirectional learning for point cloud based scene
  flow estimation.
\newblock In {\em Proc. ECCV}, 2022.

\bibitem{cordts2016cityscapes}
Marius Cordts, Mohamed Omran, Sebastian Ramos, Timo Rehfeld, Markus Enzweiler,
  Rodrigo Benenson, Uwe Franke, Stefan Roth, and Bernt Schiele.
\newblock The cityscapes dataset for semantic urban scene understanding.
\newblock In {\em Proc. CVPR}, 2016.

\bibitem{deng2009imagenet}
Jia Deng, Wei Dong, Richard Socher, Li-Jia Li, Kai Li, and Li Fei-Fei.
\newblock Imagenet: A large-scale hierarchical image database.
\newblock In {\em Proc. CVPR}, 2009.

\bibitem{dewan2016rigid}
Ayush Dewan, Tim Caselitz, Gian~Diego Tipaldi, and Wolfram Burgard.
\newblock Rigid scene flow for 3d lidar scans.
\newblock In {\em Proc. {IEEE/RSJ} Conf. on Intelligent Robots and Systems},
  2016.

\bibitem{ding2022fh}
Lihe Ding, Shaocong Dong, Tingfa Xu, Xinli Xu, Jie Wang, and Jianan Li.
\newblock Fh-net: A fast hierarchical network for scene flow estimation on
  real-world point clouds.
\newblock In {\em Proc. ECCV}, 2022.

\bibitem{dong2022exploiting}
Guanting Dong, Yueyi Zhang, Hanlin Li, Xiaoyan Sun, and Zhiwei Xiong.
\newblock Exploiting rigidity constraints for lidar scene flow estimation.
\newblock In {\em Proc. CVPR}, 2022.

\bibitem{eigen2014depth}
David Eigen, Christian Puhrsch, and Rob Fergus.
\newblock Depth map prediction from a single image using a multi-scale deep
  network.
\newblock {\em NIPS}, 2014.

\bibitem{geiger2013vision}
Andreas Geiger, Philip Lenz, Christoph Stiller, and Raquel Urtasun.
\newblock Vision meets robotics: The kitti dataset.
\newblock {\em Intl. J. of Robotics Research}, 32(11):1231--1237, 2013.

\bibitem{godard2019digging}
Cl{\'e}ment Godard, Oisin Mac~Aodha, Michael Firman, and Gabriel~J Brostow.
\newblock Digging into self-supervised monocular depth estimation.
\newblock In {\em Proc. ICCV}, 2019.

\bibitem{gonzalez2021plade}
Juan~Luis Gonzalez and Munchurl Kim.
\newblock Plade-net: towards pixel-level accuracy for self-supervised
  single-view depth estimation with neural positional encoding and distilled
  matting loss.
\newblock In {\em Proc. CVPR}, 2021.

\bibitem{gonzalezbello2020forget}
Juan~Luis GonzalezBello and Munchurl Kim.
\newblock Forget about the lidar: Self-supervised depth estimators with med
  probability volumes.
\newblock {\em NIPS}, 2020.

\bibitem{gu2019hplflownet}
Xiuye Gu, Yijie Wang, Chongruo Wu, Yong~Jae Lee, and Panqu Wang.
\newblock Hplflownet: Hierarchical permutohedral lattice flownet for scene flow
  estimation on large-scale point clouds.
\newblock In {\em Proc. CVPR}, 2019.

\bibitem{huguet2007variational}
Fr{\'e}d{\'e}ric Huguet and Fr{\'e}d{\'e}ric Devernay.
\newblock A variational method for scene flow estimation from stereo sequences.
\newblock In {\em Proc. ICCV}, 2007.

\bibitem{hur2020self}
Junhwa Hur and Stefan Roth.
\newblock Self-supervised monocular scene flow estimation.
\newblock In {\em Proc. CVPR}, 2020.

\bibitem{hur2021self}
Junhwa Hur and Stefan Roth.
\newblock Self-supervised multi-frame monocular scene flow.
\newblock In {\em Proc. CVPR}, 2021.

\bibitem{jaimez2015motion}
Mariano Jaimez, Mohamed Souiai, J{\"o}rg St{\"u}ckler, Javier Gonzalez-Jimenez,
  and Daniel Cremers.
\newblock Motion cooperation: Smooth piece-wise rigid scene flow from rgb-d
  images.
\newblock In {\em Proc. 3DV}, 2015.

\bibitem{jiang2020dipe}
Hualie Jiang, Laiyan Ding, Zhenglong Sun, and Rui Huang.
\newblock Dipe: Deeper into photometric errors for unsupervised learning of
  depth and ego-motion from monocular videos.
\newblock In {\em Proc. {IEEE/RSJ} Conf. on Intelligent Robots and Systems},
  2020.

\bibitem{jiang2019sense}
Huaizu Jiang, Deqing Sun, Varun Jampani, Zhaoyang Lv, Erik Learned-Miller, and
  Jan Kautz.
\newblock Sense: A shared encoder network for scene-flow estimation.
\newblock In {\em Proc. ICCV}, 2019.

\bibitem{jiang2022mlfvo}
Zijie Jiang, Hajime Taira, Naoyuki Miyashita, and Masatoshi Okutomi.
\newblock Self-supervised ego-motion estimation based on multi-layer fusion of
  rgb and inferred depth.
\newblock In {\em Proc. Intl. Conf. on Robotics and Automation}, 2022.

\bibitem{jiao2021effiscene}
Yang Jiao, Trac~D Tran, and Guangming Shi.
\newblock Effiscene: Efficient per-pixel rigidity inference for unsupervised
  joint learning of optical flow, depth, camera pose and motion segmentation.
\newblock In {\em Proc. CVPR}, 2021.

\bibitem{liu2019unsupervised}
Liang Liu, Guangyao Zhai, Wenlong Ye, and Yong Liu.
\newblock Unsupervised learning of scene flow estimation fusing with local
  rigidity.
\newblock In {\em Proc. IJCAI}, 2019.

\bibitem{liu2019flownet3d}
Xingyu Liu, Charles~R Qi, and Leonidas~J Guibas.
\newblock Flownet3d: Learning scene flow in 3d point clouds.
\newblock In {\em Proc. CVPR}, 2019.

\bibitem{loshchilov2017decoupled}
Ilya Loshchilov and Frank Hutter.
\newblock Decoupled weight decay regularization.
\newblock {\em arXiv preprint arXiv:1711.05101}, 2017.

\bibitem{luo2019every}
Chenxu Luo, Zhenheng Yang, Peng Wang, Yang Wang, Wei Xu, Ram Nevatia, and Alan
  Yuille.
\newblock Every pixel counts++: Joint learning of geometry and motion with 3d
  holistic understanding.
\newblock {\em IEEE PAMI}, 42(10):2624--2641, 2019.

\bibitem{lv2018learning}
Zhaoyang Lv, Kihwan Kim, Alejandro Troccoli, Deqing Sun, James~M Rehg, and Jan
  Kautz.
\newblock Learning rigidity in dynamic scenes with a moving camera for 3d
  motion field estimation.
\newblock In {\em Proc. ECCV}, 2018.

\bibitem{ma2019deep}
Wei-Chiu Ma, Shenlong Wang, Rui Hu, Yuwen Xiong, and Raquel Urtasun.
\newblock Deep rigid instance scene flow.
\newblock In {\em Proc. CVPR}, 2019.

\bibitem{mehl2023m}
Lukas Mehl, Azin Jahedi, Jenny Schmalfuss, and Andr{\'e}s Bruhn.
\newblock M-fuse: Multi-frame fusion for scene flow estimation.
\newblock In {\em Proc. WACV}, 2023.

\bibitem{meister2018unflow}
Simon Meister, Junhwa Hur, and Stefan Roth.
\newblock Unflow: Unsupervised learning of optical flow with a bidirectional
  census loss.
\newblock In {\em AAAI}, 2018.

\bibitem{menze2015object}
Moritz Menze and Andreas Geiger.
\newblock Object scene flow for autonomous vehicles.
\newblock In {\em Proc. CVPR}, 2015.

\bibitem{mur2017orb}
Raul Mur-Artal and Juan~D Tard{\'o}s.
\newblock Orb-slam2: An open-source slam system for monocular, stereo, and
  rgb-d cameras.
\newblock {\em IEEE transactions on robotics}, 2017.

\bibitem{paszke2019pytorch}
Adam Paszke, Sam Gross, Francisco Massa, Adam Lerer, James Bradbury, Gregory
  Chanan, Trevor Killeen, Zeming Lin, Natalia Gimelshein, Luca Antiga, et~al.
\newblock Pytorch: An imperative style, high-performance deep learning library.
\newblock {\em NIPS}, 2019.

\bibitem{puy2020flot}
Gilles Puy, Alexandre Boulch, and Renaud Marlet.
\newblock Flot: Scene flow on point clouds guided by optimal transport.
\newblock In {\em Proc. ECCV}, 2020.

\bibitem{Qiao2018sfnet}
Yi-Ling Qiao, Lin Gao, Yu-Kun Lai, Fang-Lue Zhang, Ming-Ze Yuan, and Shihong
  Xia.
\newblock Sf-net: Learning scene flow from rgb-d images with cnns.
\newblock In {\em Proc. BMVC.}, 2018.

\bibitem{ren2017cascaded}
Zhile Ren, Deqing Sun, Jan Kautz, and Erik Sudderth.
\newblock Cascaded scene flow prediction using semantic segmentation.
\newblock In {\em Proc. 3DV}, 2017.

\bibitem{schuster2018sceneflowfields}
Ren{\'e} Schuster, Oliver Wasenmuller, Georg Kuschk, Christian Bailer, and
  Didier Stricker.
\newblock Sceneflowfields: Dense interpolation of sparse scene flow
  correspondences.
\newblock In {\em Proc. WACV}, 2018.

\bibitem{stone2021smurf}
Austin Stone, Daniel Maurer, Alper Ayvaci, Anelia Angelova, and Rico
  Jonschkowski.
\newblock Smurf: Self-teaching multi-frame unsupervised raft with full-image
  warping.
\newblock In {\em Proc. CVPR}, 2021.

\bibitem{sun2018pwc}
Deqing Sun, Xiaodong Yang, Ming-Yu Liu, and Jan Kautz.
\newblock Pwc-net: Cnns for optical flow using pyramid, warping, and cost
  volume.
\newblock In {\em Proc. CVPR}, 2018.

\bibitem{teed2020raft}
Zachary Teed and Jia Deng.
\newblock Raft: Recurrent all-pairs field transforms for optical flow.
\newblock In {\em Proc. ECCV}, 2020.

\bibitem{teed2021raft}
Zachary Teed and Jia Deng.
\newblock Raft-3d: Scene flow using rigid-motion embeddings.
\newblock In {\em Proc. CVPR}, 2021.

\bibitem{teed2021tangent}
Zachary Teed and Jia Deng.
\newblock Tangent space backpropagation for 3d transformation groups.
\newblock In {\em Proc. CVPR}, 2021.

\bibitem{umeyama1991least}
Shinji Umeyama.
\newblock Least-squares estimation of transformation parameters between two
  point patterns.
\newblock {\em IEEE PAMI}, 13(04):376--380, 1991.

\bibitem{vedula1999three}
Sundar Vedula, Simon Baker, Peter Rander, Robert Collins, and Takeo Kanade.
\newblock Three-dimensional scene flow.
\newblock In {\em Proc. ICCV}, 1999.

\bibitem{vogel2014view}
Christoph Vogel, Stefan Roth, and Konrad Schindler.
\newblock View-consistent 3d scene flow estimation over multiple frames.
\newblock In {\em Proc. ECCV}, 2014.

\bibitem{vogel2013piecewise}
Christoph Vogel, Konrad Schindler, and Stefan Roth.
\newblock Piecewise rigid scene flow.
\newblock In {\em Proc. ICCV}, 2013.

\bibitem{vogel20153d}
Christoph Vogel, Konrad Schindler, and Stefan Roth.
\newblock 3d scene flow estimation with a piecewise rigid scene model.
\newblock {\em IJCV}, 115(1):1--28, 2015.

\bibitem{wang2022matters}
Guangming Wang, Yunzhe Hu, Zhe Liu, Yiyang Zhou, Masayoshi Tomizuka, Wei Zhan,
  and Hesheng Wang.
\newblock What matters for 3d scene flow network.
\newblock In {\em Proc. ECCV}, 2022.

\bibitem{wang2019unos}
Yang Wang, Peng Wang, Zhenheng Yang, Chenxu Luo, Yi Yang, and Wei Xu.
\newblock Unos: Unified unsupervised optical-flow and stereo-depth estimation
  by watching videos.
\newblock In {\em Proc. CVPR}, 2019.

\bibitem{wang2020flownet3d++}
Zirui Wang, Shuda Li, Henry Howard-Jenkins, Victor Prisacariu, and Min Chen.
\newblock Flownet3d++: Geometric losses for deep scene flow estimation.
\newblock In {\em Proc. WACV}, 2020.

\bibitem{wei2021pv}
Yi Wei, Ziyi Wang, Yongming Rao, Jiwen Lu, and Jie Zhou.
\newblock Pv-raft: point-voxel correlation fields for scene flow estimation of
  point clouds.
\newblock In {\em Proc. CVPR}, 2021.

\bibitem{wu2020pointpwc}
Wenxuan Wu, Zhi~Yuan Wang, Zhuwen Li, Wei Liu, and Li Fuxin.
\newblock Pointpwc-net: Cost volume on point clouds for (self-) supervised
  scene flow estimation.
\newblock In {\em Proc. ECCV}, 2020.

\bibitem{yang2020upgrading}
Gengshan Yang and Deva Ramanan.
\newblock Upgrading optical flow to 3d scene flow through optical expansion.
\newblock In {\em Proc. ECCV}, 2020.

\bibitem{yin2018geonet}
Zhichao Yin and Jianping Shi.
\newblock Geonet: Unsupervised learning of dense depth, optical flow and camera
  pose.
\newblock In {\em Proc. CVPR}, 2018.

\bibitem{zhou_diffnet}
Hang Zhou, David Greenwood, and Sarah Taylor.
\newblock Self-supervised monocular depth estimation with internal feature
  fusion.
\newblock In {\em Proc. BMVC.}, 2021.

\bibitem{zhou2022self}
Zhengming Zhou and Qiulei Dong.
\newblock Self-distilled feature aggregation for self-supervised monocular
  depth estimation.
\newblock In {\em Proc. ECCV}, 2022.

\bibitem{zou2020learning}
Yuliang Zou, Pan Ji, Quoc-Huy Tran, Jia-Bin Huang, and Manmohan Chandraker.
\newblock Learning monocular visual odometry via self-supervised long-term
  modeling.
\newblock In {\em Proc. ECCV}, 2020.

\bibitem{zou2018df}
Yuliang Zou, Zelun Luo, and Jia-Bin Huang.
\newblock Df-net: Unsupervised joint learning of depth and flow using
  cross-task consistency.
\newblock In {\em Proc. ECCV}, 2018.

\end{thebibliography}
